\pdfoutput=1
\documentclass[11pt]{article}

\usepackage[preprint]{acl}
\usepackage{booktabs}

\usepackage{times}
\usepackage{latexsym}

\usepackage[T1]{fontenc}
\usepackage[utf8]{inputenc}

\usepackage{microtype}

\usepackage{inconsolata}

\usepackage{graphicx}

\title{Measuring and Modifying the Readability of English Texts with GPT-4}

\author{Sean Trott \and Pamela D. Rivière\\
         Department of Cognitive Science, UC San Diego\\  \texttt{\{sttrott, pdrivier\}@ucsd.edu}}

\begin{document}
\maketitle
\begin{abstract}

The success of Large Language Models (LLMs) in other domains has raised the question of whether LLMs can reliably assess and manipulate the \textit{readability} of text. We approach this question empirically. First, using a published corpus of 4,724 English text excerpts, we find that readability estimates produced ``zero-shot'' from GPT-4 Turbo and GPT-4o mini exhibit relatively high correlation with human judgments ($r = 0.76$ and $r = 0.74$, respectively), out-performing estimates derived from traditional readability formulas and various psycholinguistic indices. Then, in a pre-registered human experiment ($N = 59$), we ask whether Turbo can reliably make text easier or harder to read. We find evidence to support this hypothesis, though considerable variance in human judgments remains unexplained. We conclude by discussing the limitations of this approach, including limited scope, as well as the validity of the ``readability'' construct and its dependence on context, audience, and goal.

\end{abstract}

\section{Introduction}\label{sec:intro}

The ease with which a text can be read or understood is called \textit{readability}. Measuring and modifying readability has been a topic of interest for decades \cite{lively1923method, flesch1948new, crossley2023large}, with potential applications ranging from selecting and curating educational materials \cite{solnyshkina2017evaluating, creutz-2024-correcting, liu-lee-2023-hybrid} to making legal, medical, or other technical documents more accessible \cite{ghosh-etal-2022-finrad, rosati-2023-grasum, chen-etal-2023-ncuee}. Methods for \textit{assessing} readability, in turn, include: tests of reading comprehension, formulas incorporating basic text features \cite{lively1923method, flesch1948new} or psycholinguistic variables \cite{kyle2015automatically}, and approaches using supervised learning to estimate readability from labeled text data \cite{schwarm2005reading, martinc2021supervised}. 

Recent advances in Large Language Models (LLMs) \cite{brown2020language} has led to interest in exploring the capacities and applications of these systems---including measuring and modifying the readability of text \cite{ribeiro-etal-2023-generating, li2023large, crossley2023using, patel2023improving, farajidizaji2023possible}. In the current work, we approach this question empirically. 

In Section \ref{sec:related}, we describe in more detail past work on measuring and modifying readability of text automatically. Then, in Section \ref{sec:study1}, we empirically assess the ability of a state-of-the-art LLM (GPT-4 Turbo) to measure readability ``zero-shot''. Next, in a pre-registered human experiment, we ask whether GPT-4 Turbo can be used to modify text readability (Section \ref{sec:study2}). Finally, we conclude by discussing the implications of the current work (Section \ref{sec:discussion}), as well as its limitations (Section \ref{sec:limitations}). Note that all code and data can be found on GitHub: \url{https://github.com/seantrott/llm_readability}.

\section{Related Work}\label{sec:related}

As described in Section \ref{sec:intro}, efforts to quantify the readability of text date back at least a century \cite{lively1923method}. For many decades, approaches relied on hand-crafted features thought to correlate with (or be causally implicated in) text readability, such as the average length of words or sentences \cite{flesch1948new}. As \citet{vajjala-2022-trends} describe, dominant approaches have gradually shifted towards treating readability assessment as a supervised machine learning problem, i.e., training a system to produce representations that facilitate the prediction of ``gold standard'' human readability judgments---though researchers continue to test the viability of hand-crafted features as an alternative or complementary approach \cite{deutsch-etal-2020-linguistic, wilkens-etal-2024-exploring}. Pre-trained language models seem potentially well-suited to this task; indeed, past work \cite{crossley2023large} suggests that fine-tuning these models can produce estimates that align with human readability judgments.

\textit{Modifying} readability is also of considerable interest, with most research focusing on making text easier to read, e.g., for journal abstracts \cite{li2023large} or math assessments \cite{patel2023improving}. \citet{cardon-bibal-2023-operations} provide a useful overview of the distinct \textit{operations} used in Automatic Text Simplification (ATS), including splitting up long sentences \cite{nomoto-2023-issues} and simplifying or substituting individual words \cite{paetzold-specia-2017-lexical}. As with work on measuring readability, this research has gradually shifted from explicit, rule-based approaches to systems that ``learn'' appropriate transformations using an annotated corpus \cite{cardon-bibal-2023-operations}, sometimes tailored with psycholinguistic features \cite{qiao-etal-2022-psycho}.

Recent research has used \textit{prompt engineering} approaches to ask whether Large Language Models (LLMs) can modify text \cite{farajidizaji2023possible, ribeiro-etal-2023-generating, liu2023benchmarking, creutz-2024-correcting, imperial-tayyar-madabushi-2023-flesch, pu-demberg-2023-chatgpt, luo-etal-2022-readability, kew-etal-2023-bless}, with some studies asking whether text can be modified to some \textit{target readability level}, e.g., a target Flesch score \cite{flesch1948new}. Even with ``zero-shot'' prompting (i.e., no examples provided), LLMs appear to be surprisingly successful at modifying text readability in the desired direction---though not necessarily to the desired text level \cite{liu2023benchmarking}. In some cases, a residual correlation is found between the readability of the original text and the modified text \cite{farajidizaji2023possible}.

\section{Study 1: Measuring Readability}\label{sec:study1}

In Study 1, we focused on the ability of LLMs to estimate the readability of text excerpts ``zero-shot'' (i.e., without any labeled examples in the prompt). We asked: given a corpus of human readability estimates \cite{crossley2023large}, how well can an LLM equipped solely with instructions and a definition of readability produce outputs that correlate reliably with human judgments? 

\subsection{CLEAR Dataset}

We used the CommonLit Ease of Readability (CLEAR) Corpus \cite{crossley2023large}, which contains human estimates of readability for 4,724 text excerpts. The CLEAR Corpus was created by \citet{crossley2023large} by sampling text excerpts (between 140-200 words) from various databases (e.g., Project Gutenberg). It includes fiction and non-fiction, and spans a range from 1875 to 2020. Excerpts were then normed by asking a sample of teachers to rate pairs of items for their relative readability. These pairwise judgments were then aggregated to create a readability index for each individual passage.

\subsection{Models}

Our primary goal was assessing the reliability of using a state-of-the-art LLM in estimating readability. To this end, we used two state-of-the-art proprietary OpenAI models: GPT-4 Turbo and GPT-4o mini. We accessed both models using the OpenAI Python API: Turbo (\textsc{gpt-4-1106-preview}) and 4o mini (\textsc{gpt-4o-mini-2024-07-18}). Because both models are closed-source, it is unclear how many parameters each model has or how much data it was trained on.

\subsection{Zero-shot Annotation Procedure}

Both OpenAI models were provided with the same system prompt meant to approximate the context of participants in the original CLEAR corpus \cite{crossley2023large} (``You are an experienced teacher, skilled at identifying the readability of different texts.''). Each text excerpt was presented to the model in a separate prompt (i.e., rather than in succession), along with instructions explaining that the goal was to rate the excerpt for how easy it was to read and understand, on a scale from 1 (very challenging to understand) to 100 (very easy to understand); the exact instructions can be found in Appendix \ref{sec:appendix_instructions}. Each models' responses were produced using a temperature of 0, with a maximum number of tokens of 3. Response strings were then converted to numeric values in Python.

\subsection{Results}

\begin{figure}
    \centering
    \includegraphics[width=0.75\linewidth]{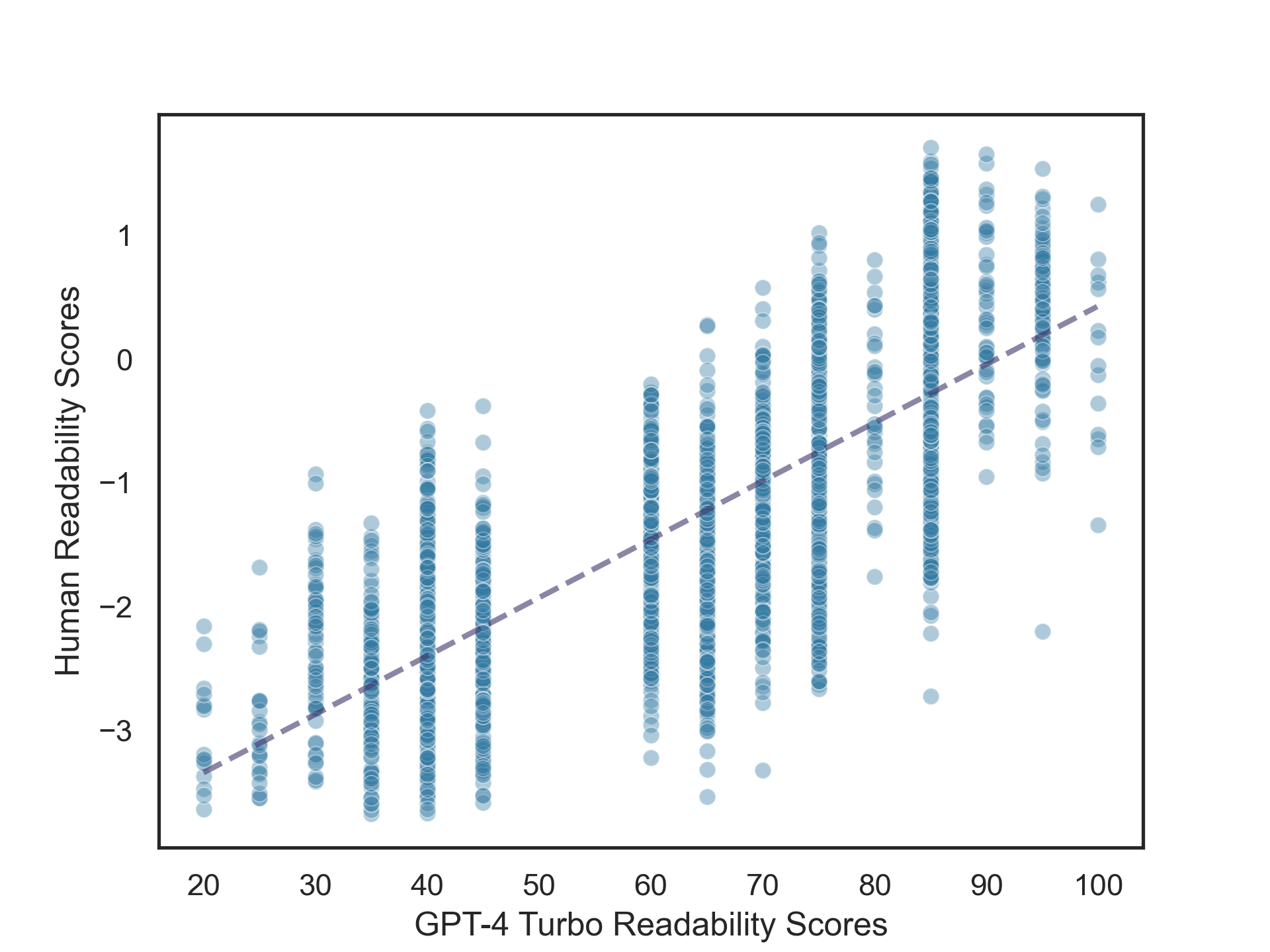}
    \caption{Relationship between ratings elicited by GPT-4 Turbo and average human readability judgments ($R^2 = 0.58$).}\label{fig:study1_main_result}
\end{figure}

We first asked how well ratings from GPT-4 Turbo and GPT-4o mini predicted human readability scores from the CLEAR dataset \cite{crossley2023large}. Concretely, this was operationalized by asking to what extent LLM-generated ratings correlated with human ratings. We found that ratings from each model were positively correlated with human readability: Turbo ($r = 0.76$) and GPT-4o mini ($r = 0.74$); see also Figure \ref{fig:study1_main_result} for the Turbo results specifically.\footnote{Ratings between Turbo and 4o were also highly correlated ($r = 0.81$).} For comparison, the correlation between two random splits within the CLEAR corpus was only $r = 0.63$.

In terms of predictive power, these correlation metrics would correspond to an $R^2$ of $.54$ (for 4o mini) or $.577$ (for Turbo). We compared this predictive power to several psycholinguistic variables known to correlate with about readability \cite{kyle2018tool}: log word frequency \cite{brysbaert2009moving}, word concreteness \cite{brysbaert2014concreteness}, and word age of acquisition \cite{kuperman2012age}. For each variable, we calculated the \textit{average} across all words in a given passage that occurred in the relevant dataset. A linear model including all three psycholinguistic predictors explained approximately $36\%$ of the variance in human readability judgments ($R^2 = 0.36$). Each variable was significantly related: frequency $[\beta = 0.82, SE = 0.13, p < .001]$, concreteness $[\beta = 1.76, SE = 0.11, p < .001]$, and age of acquisition $[\beta = -0.56, SE = 0.06, p < .001]$. Thus, psycholinguistic properties of words in a passage are useful for predicting readability judgments, but under-perform ratings elicited from GPT-4 Turbo and GPT-4o mini.\footnote{Of course, taking the average of these variables across an entire passage is a relatively coarse measure and likely represents a \textit{lower-bound} on their predictive efficacy.}

\begin{figure}
    \centering
    \includegraphics[width=0.75\linewidth]{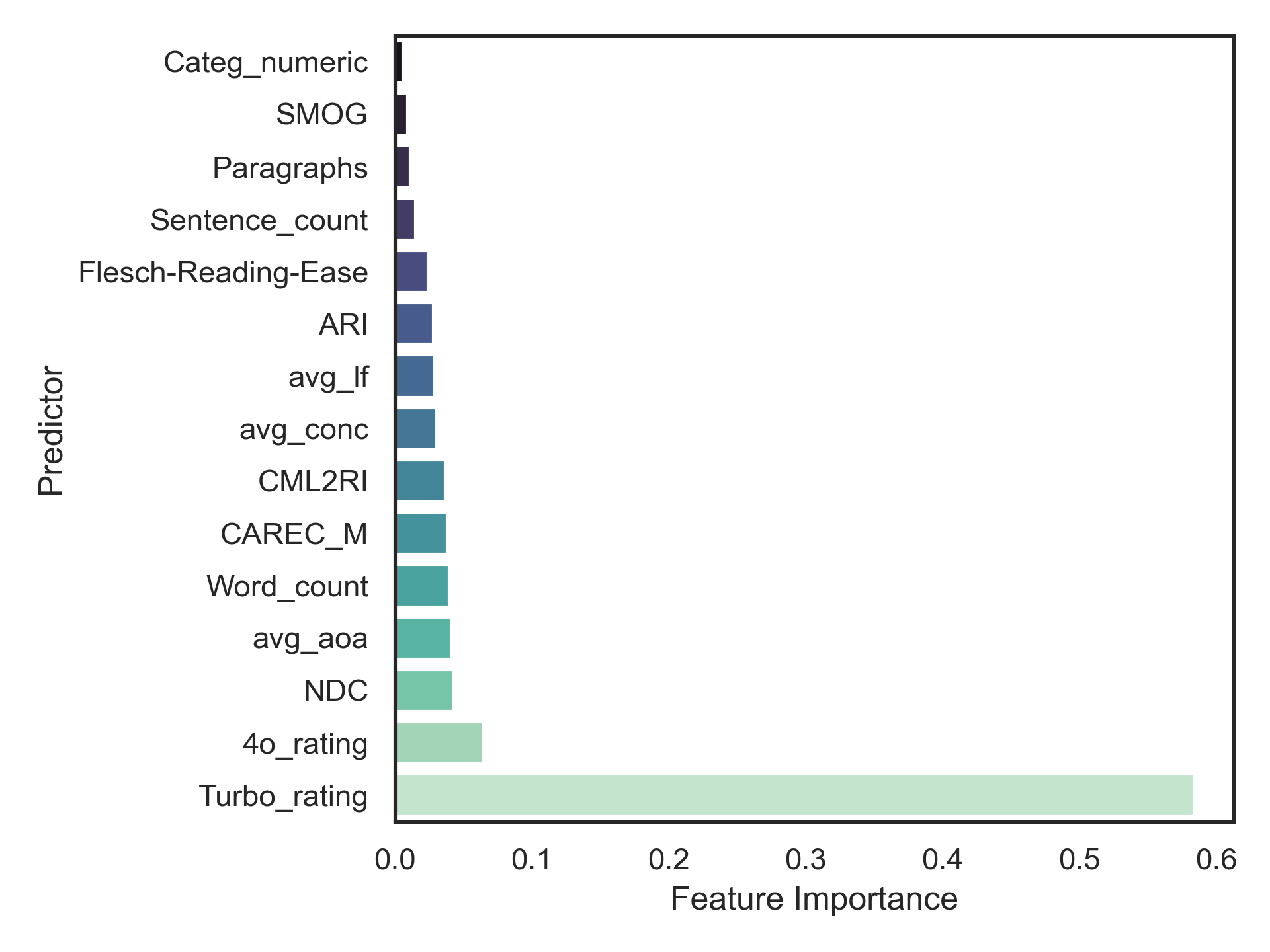}
    \caption{Feature importance scores for each predictor, as determined using a random forest regression. A higher value indicates that this feature was more useful for predicting human readability judgments.}\label{fig:rf_importances}
\end{figure}

As a final test of predictive power, we entered the metrics considered above---along with measures like the number of words and sentences, and estimates derived from traditional readability formulas---as predictors in a random forest regression and compared their \textit{feature importance scores}.\footnote{No maximum depth was used, and the random state was set to $0$.} These scores can be interpreted as reflecting the extent to which the inclusion of a particular feature (e.g., ratings from Turbo) reduce prediction error when predicting human readability. All measures were $z$-scored before fitting the model. As depicted in Figure \ref{fig:rf_importances}, Turbo's ratings were assigned the highest importance, followed by the average age of acquisition scores.

\section{Study 2: Modifying Readability}\label{sec:study2}

In Study 2, we asked whether a state-of-the-art LLM  could successfully \textit{modify} (as opposed to simply \textit{measure}) the readability of texts. GPT-4 Turbo performed best in Study 1, so we selected Turbo for modifying text readability as well. We approached this question in the following way: given instructions to make a text excerpt \textit{easier} or \textit{harder}, can an LLM produce a modified version that an independent pool of human judges rate as easier or harder than the original? Although it is unlikely that making texts \textit{harder} to read is a desirable goal, we included this condition as a control (i.e., to ensure that modified passages were not always rated as easier to read). This study was pre-registered on the Open Science Framework (OSF).\footnote{Link to pre-registration for text modification: \url{https://osf.io/vtwug}. Link to pre-registration for human experiment: \url{https://osf.io/6hmej}.}

\subsection{Materials}

To make this question empirically tractable, we selected a random sample of $100$ excerpts from the original CLEAR corpus. Each excerpt was then presented to GPT-4 Turbo twice, with two different sets of instructions asking Turbo to make the excerpt easier or harder to read (exact prompting and instructions found in Appendix \ref{sec:appendix_instructions}). As in Study 1, Turbo was first provided with a system prompt (``You are an experienced writer, skilled at rewriting texts.''); a temperature of 0 was used, and the maximum number of tokens was set to the number of tokens in the original excerpt, plus a ``buffer'' of 5 tokens. Additionally, we specified that the modified version should be of approximately the same length as the original.

This resulted in $300$ items altogether. For the human study, these items were assigned to $6$ lists using a Latin Square design, where each list had approximately $50$ items. No list contained multiple versions of the same item. Note that in some cases, the modified version produced by Turbo cut-off in mid-sentence; we further modified these excerpts by removing the final sentence fragment. The experiment was designed on the Gorilla experimental design platform \cite{anwyl2018gorillas}.

\subsection{Participants}

Our target $N$ was $60$ participants ($10$ per list). We anticipated a non-zero exclusion rate, so we intended to recruit $70$ participants via Prolific; due to an error in the recruiting platform, we recruited only $69$. As per our pre-registration, we excluded participants whose readability ratings for the \textit{original} text excerpts exhibited a correlation with the gold standard of $r < .1$; this resulted in the removal of $10$ participants. Participants were paid $\$6.00$ and the median completion time was $34$ minutes and $21$ seconds (an average rate of $\$10.48$ per hour). In the final pool of participants, $34$ participants identified as female ($22$ male, $2$ non-binary, and $1$ preferred not to answer); the average self-reported age was $40.77$ (SD = $14$). Note that unlike the CLEAR corpus \cite{crossley2023large}, we did not recruit specifically teachers or other employees in the education sector.

\subsection{Procedure}

Each participant rated the readability of a series of $50$ text excerpts on a scale from 1 (very challenging to understand) to 5 (very easy to understand). Participants were instructed to consider factors such as ``sentence structure, vocabulary complexity, and overall clarity''; they were also reminded to try to focus on the readability of the passage itself, as opposed to the complexity of the topic. No participant rated multiple versions of the same item and the order of items was randomized across trials.

\subsection{Results}

We carried out three pre-registered analyses in R using the \textit{lme4} package \cite{bates2011mixed}. In the case of fitting mixed effects models, we began with maximal random effects structure and reduced as needed for model convergence \cite{barr2013random}. Nested model comparisons were conducted by comparing a full model to a reduced model omitting only the variable of interest, using a log-likelihood ratio test (LRT). 

\begin{figure}
    \centering
    \includegraphics[width=0.75\linewidth]{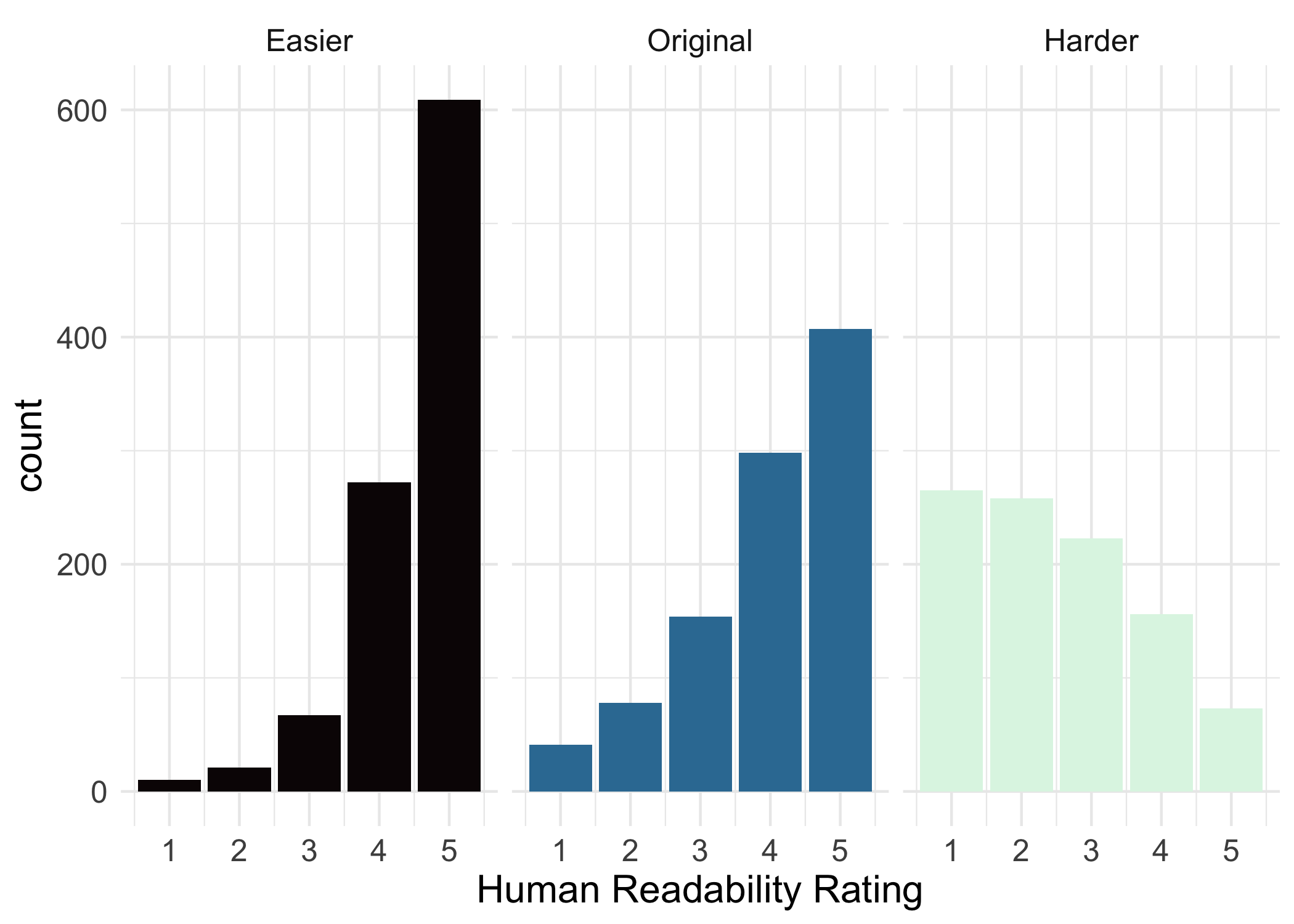}
    \caption{Distribution of human readability judgments for each text condition.}\label{fig:study2_histogram}
\end{figure}

Human readability judgments were predicted by the contrast between \textit{Easy} and \textit{Hard} $[\chi^2(1) = 97.58, p < .001]$, between \textit{Easy} and \textit{Original} $[\chi^2(1) = 32.4, p < .001]$, and between \textit{Hard} and \textit{Original} $[\chi^2(1) = 74.75, p < .001]$. That is, significant variance in human readability judgments was explained by the condition under which a particular passage was produced. As depicted in Figure \ref{fig:study2_histogram}, excerpts in the \textit{Easier} condition were rated as the most readable ($M = 4.48, SD = 0.8$), excerpts in the \textit{Harder} condition were rated as the least readable ($M = 2.5, SD = 1.25$), with excerpts in the \textit{Original} condition between the two ($M = 3.97, SD = 1.13$).

\section{Discussion}\label{sec:discussion}

Our primary question was whether state-of-the-art LLMs could be used to \textit{measure} and \textit{modify} the readability of a text excerpt. In Study 1, we found that ratings from GPT-4 and GPT-4o mini ratings were strongly correlated with gold standard ratings, though Turbo's ratings ($r = 0.76$) were slightly more correlated than ratings from GPT-4o mini; consistent with other recent work using LLMs for text annotation \cite{trott2024can, trott2024large}, this correlation was higher than the correlation between random splits of human ratings \cite{cross-etal-2023-glossy}. Further, Turbo's ratings were the best predictor of human readability judgments of all the variables tested (see Study \ref{sec:study1}), including several psycholinguistic variables and other readability formula estimates.

In Study 2, we asked Turbo to produce easier or harder versions of $100$ sample excerpts from the same corpus \cite{crossley2023large}. In a pre-registered human study, participants consistently rated the \textit{easier} versions as easier to read, and the \textit{harder} versions as harder to read---though notably, there was a correlation between the readability of the original text passage and the modified passage (see Figure \ref{fig_appendix:flesch_comparison_og}).

As with other recent work \cite{farajidizaji2023possible, liu2023benchmarking, ribeiro-etal-2023-generating}, these results provide a proof-of-concept that LLMs may be useful for both measuring and modifying text readability, at least as operationalized here. Unlike past work \cite{ribeiro-etal-2023-generating, farajidizaji2023possible}, we do not investigate the question of modification to \textit{target readability levels}, though we do collect novel human judgments to validate the success of GPT-4 Turbo's modifications (Study \ref{sec:study2}). Of course, considerable open questions about the viability of this approach remain. These questions are all explored in more detail in the Limitations section below.

\section{Limitations}\label{sec:limitations}

One limitation, particularly of Study 2, is scope: because we planned to collect human annotations for each excerpt, we considered only $100$ text excerpts, and compared the performance of only one model (GPT-4 Turbo). The results of this study can be seen as a proof-of-concept, which future work can build on with larger samples and more sophisticated prompt engineering techniques. More generally, a limitation of both studies is that they considered excerpts from a single readability dataset only. Future work ocould explore other readability datasets and benchmarks to ask how well these results generalize. Relatedly, we aimed to use a prompt that would allow fair comparisons to data collected from humans and thus did not explore alternative prompt engineering techniques. However, future work could explore how prompting strategies affect model performance and behavior.

A further limitation of Study 2 is that we did not assess the modified excerpts in terms of their faithfulness to the original text. Evaluating the quality of summaries is notoriously difficult \cite{wang2019task}, though recent work \cite{liu2023benchmarking} has made use of automated metrics like BERTScore \cite{Zhang*2020BERTScore:}. Future work would benefit from another human study that asks directly about the \textit{quality} of the modified texts; these results could then be used to validate automated metrics. Relatedly, the evaluators in Study 2 did not have particular experience in assessing readability or linguistics more generally; future work could recruit annotators with more expertise to create more nuanced readability ratings.

A final limitation is the question of what the \textit{construct} of readability means in the first place, and how best to measure it. Construct validity is by no means a new challenge for work in NLP generally \cite{raji2021ai} or readability specifically \cite{crossley2008assessing}. ``Readability'' may not be a unitary construct; different stakeholders may construe readability in different ways depending on their goal (e.g., making a product manual accessible vs. curating educational materials) and audience (e.g., school-aged children vs. professionals). Further, different formulas or automated metrics emphasize different properties of a text, making implicit or explicit assumptions about the underlying construct. The current work relied on human judgments of readability as a ``gold standard'', using both existing corpora \cite{crossley2023large} and novel data (Study 2). By these metrics, using Turbo to measure and modify readability was modestly successful. Yet it is unclear whether these results generalize to other texts, contexts, goals, or audiences. Thus, future work could benefit from additional research on ``benchmarking'' readability itself \cite{kew-etal-2023-bless} and whether different benchmarks are needed for different senses of readability.

\section{Lay Summary}

We asked whether Large Language Models (LLMs) were able to measure---and later change---the ``readability'' of snippets of English-language text taken from the openly available CLEAR Corpus. We presented text excerpts to GPT-4o mini and GPT-4 Turbo, collected their readability ratings on a scale from 1 (very difficult) to 100 (very easy), and found that their ratings were positively correlated with the corresponding human readability judgments. Notably, GPT-4 Turbo outperformed readability estimates from -4o mini, and from a battery of more traditional readability measures. We next instructed GPT-4 Turbo---the best-performing model---to rewrite each text excerpt to make it “easier” or “harder” to read relative to the original, while keeping the length of the rewritten excerpts roughly the same as the original. We then conducted a validation study to determine whether human judges found the rewritten excerpts easier or harder to read. Human judges produced readability ratings for each rewritten text excerpt between 1 (difficult) to 5 (easy). When GPT-4-Turbo rewrote a text to read more easily, human judges did in fact find it easier to read than texts rewritten to seem harder to read. This suggests that off-the-shelf LLMs are capable of assessing text readability, and can modify readability to (coarsely defined) target levels.  

\section{Ethical Considerations}

All data collected from human participants has been fully anonymized before analysis or publication.

One potential risk with research on automatic text simplification is that tools will be deployed in various applied settings (e.g., education) before they are ready. As we discussed in the Limitations section (Section \ref{sec:limitations}), we believe there are a number of open questions remaining with this kind of research and do not intend for these results to signal that LLMs could and should be used for measuring and modifying readability in an applied domain at this time.

\section*{Acknowledgments}

We are grateful to advice and suggestions from Benjamin Bergen and Cameron Jones. Pamela Rivi\`ere was funded by the Chancellor's Postdoctoral Fellowship Program at UC San Diego. 

\bibliography{anthology, anthology2, custom}

\appendix

\section{Appendix}
\label{sec:appendix}

\subsection{Instructions for Study 1 and Study 2}\label{sec:appendix_instructions}

In this section, we report the exact prompts used to elicit readability judgments from GPT-4 Turbo. Note that symbols like ``{EXCERPT}'' indicate that the text of the excerpt was inserted in this section of the prompt.

\textbf{Study 1 Instructions}:

\begin{quote}
Read the text below. Then, indicate the readability of the text, on a scale from 1 (extremely challenging to understand) to 100 (very easy to read and understand). In your assessment, consider factors such as sentence structure, vocabulary complexity, and overall clarity. 

<Text>:{EXCERPT}</Text>

On a scale from 1 (extremely challenging to understand) to 100 (very easy to read and understand), how readable is this text?. Please answer with a single number.
\end{quote}

\textbf{Study 2 Instructions}:

\begin{quote}

    Read the passage below. Then, rewrite the passage so that it is easier/harder to read.
    
    When making the passage more/less readable, consider factors such as sentence structure, vocabulary complexity, and overall clarity. However, make sure that the passage conveys the same content.
    
    Finally, try to make the new version approximately the same length as the original version.

    <Text>:{EXCERPT}</Text>
    
    As described in the instructions, please make this passage easier/harder to read, while keeping the length the same.
\end{quote}

\subsection{Exploratory Analyses for Study 1}\label{sec:study1_appendix}

\begin{figure}
    \centering
    \includegraphics[width=0.95\linewidth]{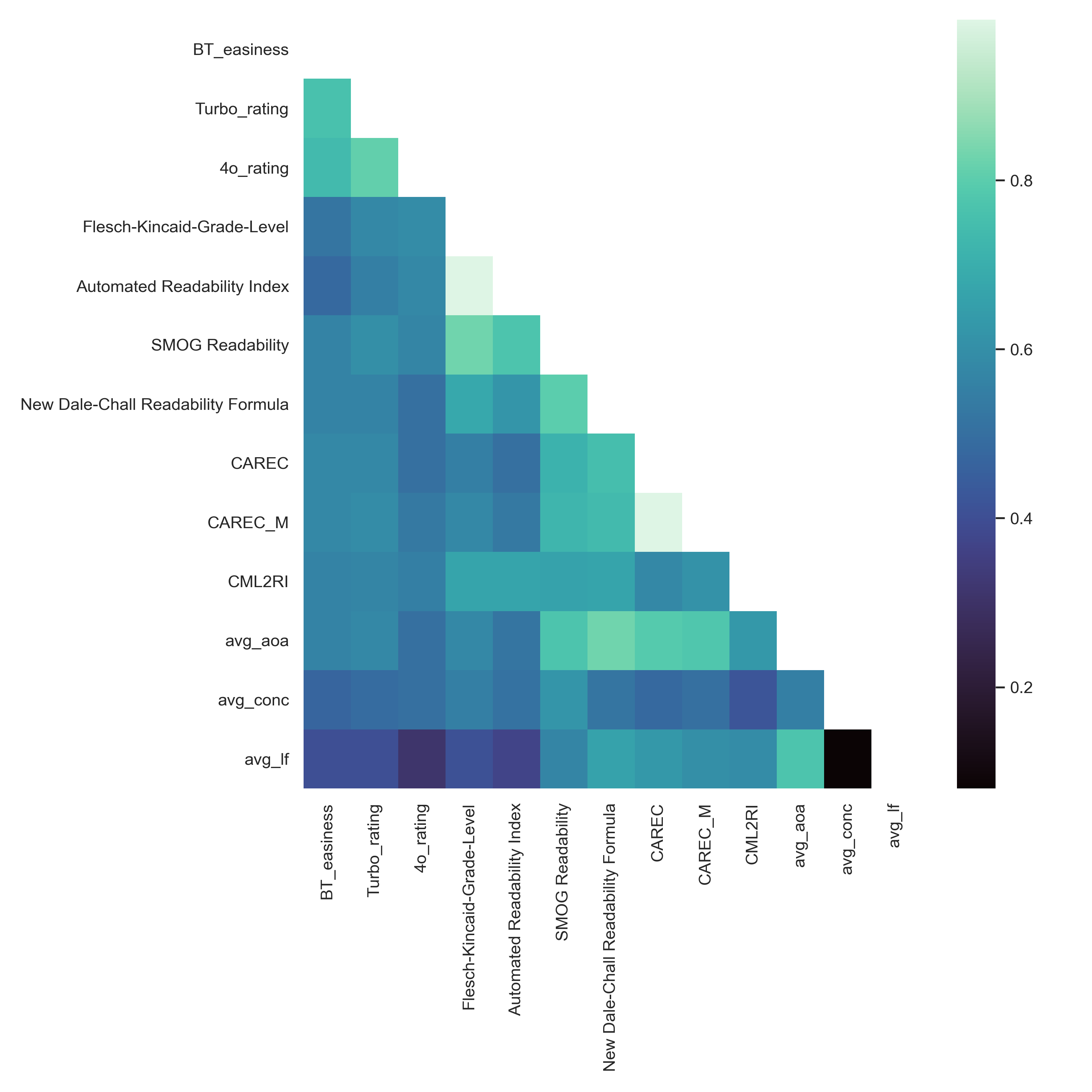}
    \caption{Correlation matrix between all the variables considered in Study 1. Correlation coefficients have all been transformed to absolute values for easier comparison.}\label{fig:corr_matrix}
\end{figure}

We also constructed a correlation matrix of all the variables considered: see Figure \ref{fig:corr_matrix}).

\subsection{Exploratory Analyses for Study 2}

\begin{figure}
    \centering
    \includegraphics[width=0.75\linewidth]{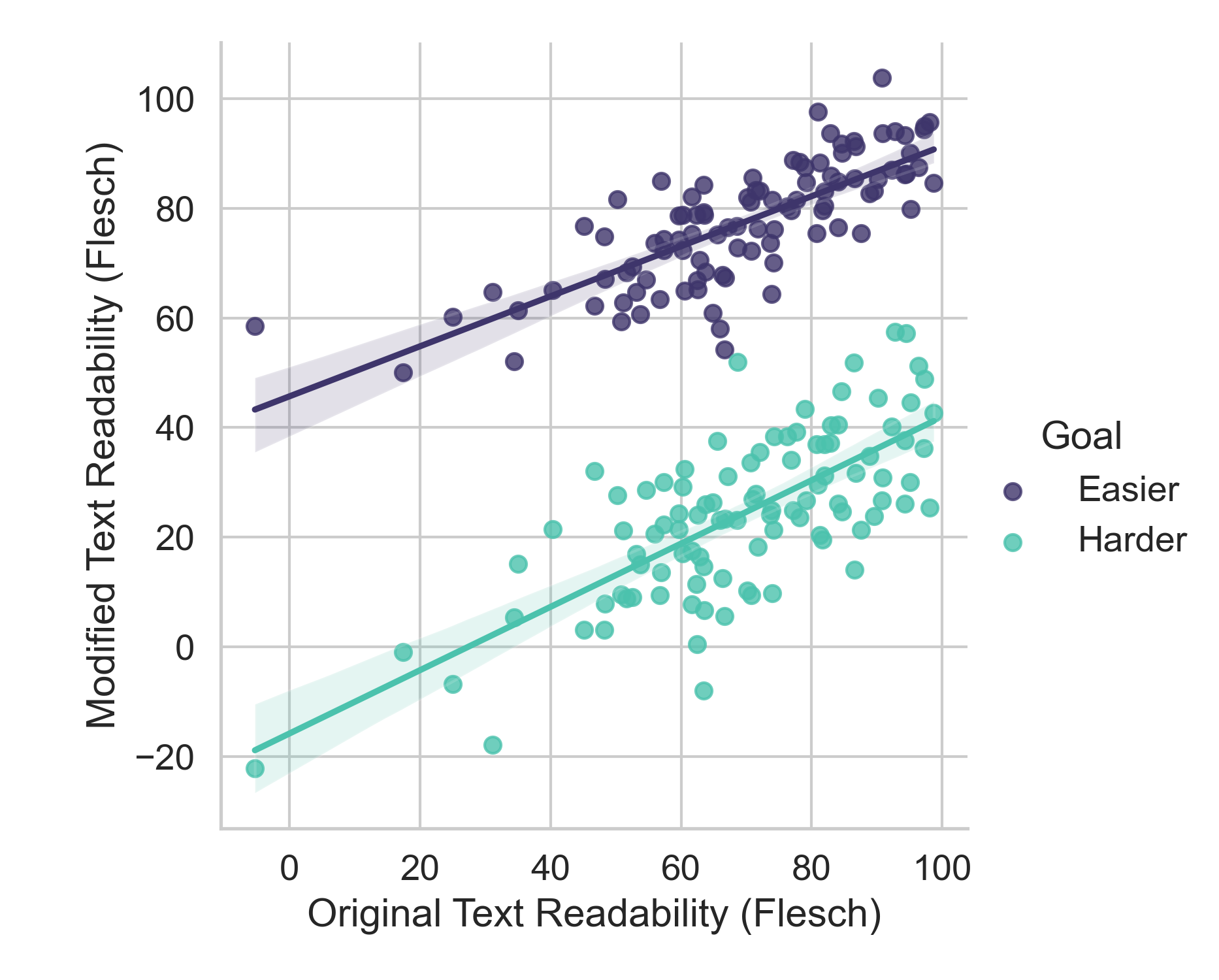}
    \caption{Comparison of Flesch readability for the original version and modified version, according to Turbo's instructions.}\label{fig_appendix:flesch_comparison_og}
\end{figure}

In an exploratory analysis, we asked whether the readability of the original text excerpt was correlated with the readability of the modified version. Consistent with \cite{farajidizaji2023possible}, we found a positive correlation: that is, Turbo successfully modified texts to be easier or harder to read, depending on the instructions, but the readability of the modified text exhibited a residual correlation with the original text's readability (see Figure \ref{fig_appendix:flesch_comparison_og}).

\begin{figure}
    \centering
    \includegraphics[width=1\linewidth]{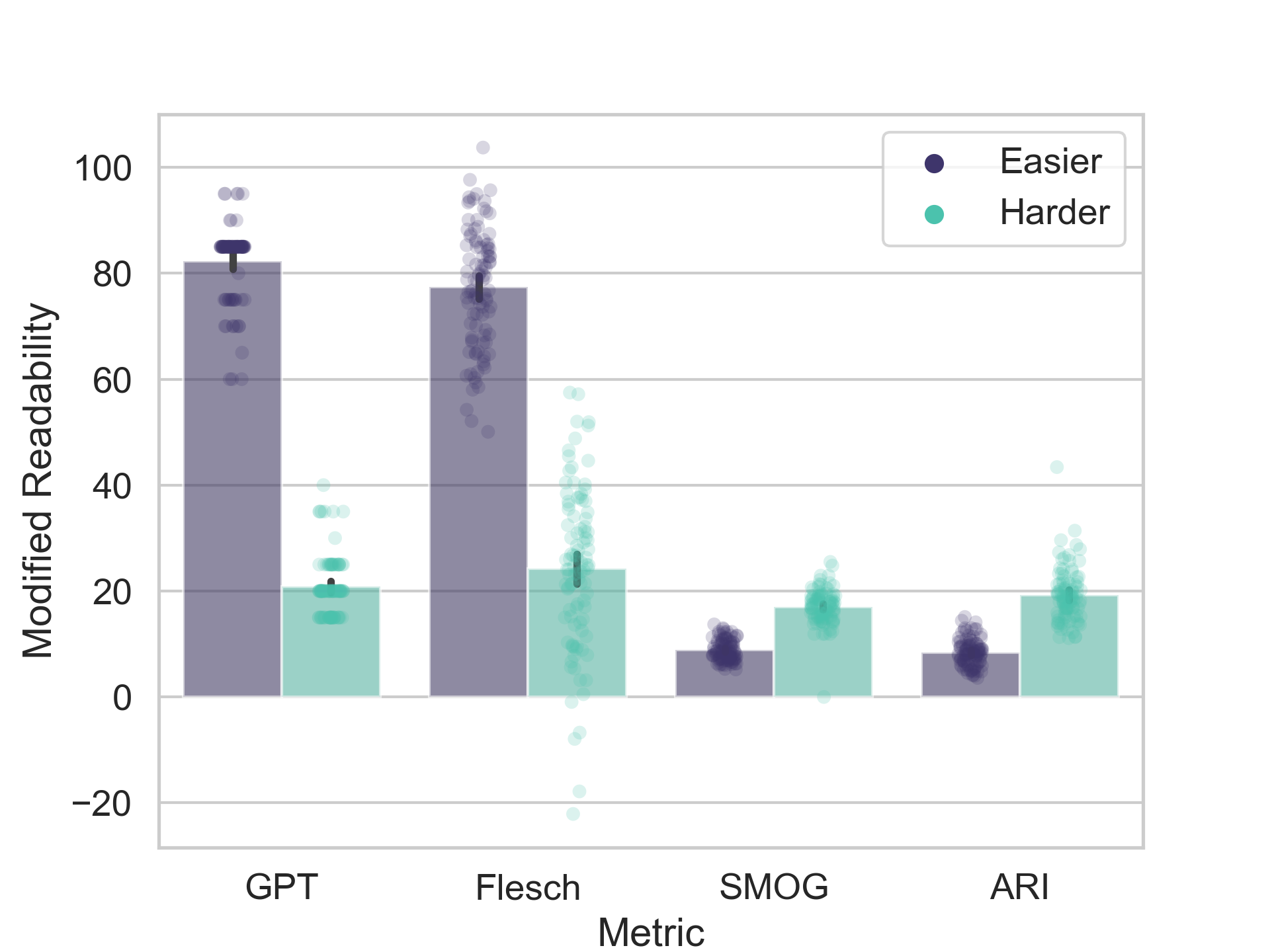}
    \caption{Comparison of automated readability scores for the modified text excerpts.}\label{fig_appendix:all_metrics_study2}
\end{figure}

Additionally, we calculated the readability of the modified texts using automated readability formulas, e.g., the Flesch Reading Score \cite{flesch1948new}. We then asked whether the modified versions varied in the expected direction along each metric in question, according to whether Turbo was instructed to make the text easier or harder to read. We found that the modified versions varied in the expected direction according to automated readability metrics as well (see Figure \ref{fig_appendix:all_metrics_study2}).

\end{document}